\documentclass{article}
\usepackage{geometry}
\usepackage{url}
\usepackage{authblk}
\usepackage{siunitx}
\usepackage{booktabs}
\usepackage[utf8]{inputenc}
\usepackage{multirow}
\usepackage[pdftex]{graphicx}
\usepackage[linesnumbered,ruled,vlined]{algorithm2e}
\usepackage{rotating}
\usepackage[extra]{tipa}
\usepackage{xcolor}
\newtheorem{definition}{Definition}
\usepackage{amssymb}
\usepackage{amsmath}
\usepackage{amsfonts}

\newcommand\RotText[1]{\fontsize{9}{9}\selectfont
  \rotatebox[origin=c]{60}{\parbox{2.6cm}{%
\centering#1}}}

\title{Whom to Test? Active Sampling Strategies for Managing COVID-19}
\author{ Yingfei Wang\thanks{All authors have contributed equally.}\\
  Foster School of Business\\
  University of Washington\\
  \texttt{yingfei@uw.edu} \\
    \and
    Inbal Yahav \\
    Coller School of Management \\
    Tel Aviv Univeristy \\
    \texttt{inbalyahav@tauex.tau.ac.il} \\
    
    \and
    Balaji Padmanabhan \\
    Muma College of Business\\
    University of South Florida\\
    \texttt{bp@usf.edu}
    }
\date{December 20, 2020}

\begin{document}

\maketitle
\pagebreak
\begin{center}
\vspace*{1cm}
\large
\textbf{Whom to Test? Active Sampling Strategies for Managing COVID-19}
\vspace{0.5cm}
\end{center}

\begin{abstract}
    This paper presents methods to choose individuals to test for infection during a pandemic such as COVID-19, characterized by high contagion and presence of asymptomatic carriers. The smart-testing ideas presented here are motivated by active learning and multi-armed bandit techniques in machine learning. Our active sampling method works in conjunction with quarantine policies, can handle different objectives, is dynamic and adaptive in the sense that it continually adapts to changes in real-time data. The bandit algorithm uses contact tracing, location-based sampling and random sampling in order to select specific individuals to test. Using a data-driven agent-based model simulating New York City we show that the algorithm samples individuals to test in a manner that rapidly traces infected individuals. Experiments also suggest that smart-testing can significantly reduce the death rates as compared to current methods such as testing symptomatic individuals with or without contact tracing.
\end{abstract}

\section{Introduction}\label{sec:intro}
This paper presents a method to actively sample individuals in a population as a way to mitigate the spread of pandemics such as COVID-19. Sampling algorithms are commonly used in machine learning to acquire training data labels for classification ("active learning" \cite{AL-MJ, pz2006}) and in bandit algorithms \cite{bandit-wolpert} to explore complex search spaces through exploration and exploitation. The method we present in this paper builds on these ideas, but do so in the context of containing the spread of an epidemic in a population. 

The literature on managing disease spread through ideas based on active sampling is primarily from the public health area. There are two  reasons why this literature has considered sampling, although both these are sometimes intertwined in the context of population surveillance \cite{lee2010principles}. 

The first is to estimate the actual incidence or spread of a disease, such as HIV, in a population. In the case of estimating HIV incidence , methods such as a population survey and "sentinel surveillance" \cite{magnani2005review, fylkesnes_studying_1998} have been shown to be useful. The population survey in a catchment area (i.e. where there is likely disease) is (stratified/cluster) random sampling, and is generally considered the gold standard but also known to be expensive \cite{lee2010principles}. Sentinel surveillance, in contrast, gathers data from a subset of venues (the "sentinels", e.g. clinics) where the subset with the disease is likely to visit for treatment. Recognizing that sensitive diseases like HIV may be prevalent in hard-to-reach populations who might stay under the radar due to activities that are illegal or illicit (e.g. sex workers, drug users) the literature \cite{magnani2005review, goel2010assessing} has critically studied other techniques such as snowball sampling (starting with a few seeds who have the disease and asking them to name others who might), respondent-driven sampling (similar to snowball, but longer chains with fewer referrals per seed) and time-location sampling (e.g. sampling users at nights in districts with brothels). With sensor data newer forms of sampling are emerging as well. It has been shown for instance, that to quantify disease spread in livestock that sampling the nodes (physical locations) where there is greatest movement can be effective \cite{dawson2015sampling}. 

The second reason for considering sampling strategies is to mitigate the spread of disease by proactively identifying individuals who need to be quarantined or tested. One method to do this, common in the public health literature is contact-tracing \cite{PhysRevE.66.056115, eames2003contact, underwood_contact_2003}. The idea is to sample contacts of known infected individuals for presence of disease, and, along with quarantine, is the main mechanism used worldwide during the COVID-19 pandemic \cite{salathe2020covid, hellewell2020feasibility}. Traditionally contact-tracing is done through questionnaires, but newer automated contact tracing based on tracking cell phone proximities have also been proposed \cite{ferretti2020quantifying}. 

The SARS-Cov-2 pathogen and its transmission have the following unique characteristics, though, that make none of these sampling methodologies sufficient by itself.
\begin{itemize}
    \item The possibility of pre-symptomatic and asymptomatic spread rule out pure sentinel-based strategies that would test those who report at clinics with symptoms. For the same reason,  contact tracing alone would be insufficient. This is because contact-tracing pursues only contacts of those known to have tested positive. Asymptomatic carriers may have also been transmitting the disease to their contacts, but those will not be actively sought by this strategy. Hence, random sampling is also useful.
    \item The pathogen has also been shown to survive on surfaces or in air in specific locations for extended periods of time. Hence, location-based sampling may also be useful.
    \item There is uncertainty about whether individuals develop long-lasting immunity. Hence sampling strategies may have to allow for re-testing the same individuals repeatedly.
    \item Information about how covariates influence getting the disease, and being affected by it, are still evolving. For instance, while we know that comorbities and age affect outcomes, there is still uncertainty about whether this knowledge is complete. Moreover, how covariates might affect contracting the disease is still uncertain. In this environment, covariate-based sampling strategies alone might be insufficient. Sampling strategies that target specific \emph{individuals} to test is therefore critical. 
    \item The highly infectious nature of the pathogen and the extent of worldwide spread have placed constraints on testing capacities, requiring prioritizing tests if they are to be useful in mitigating the spread. 
\end{itemize}

The method presented in this paper addresses all five aspects noted above. This paper presents dynamic active sampling strategies based on multi-armed bandit algorithms that can optimally combine the different sampling ideas to generate real-time lists of whom to sample. Bandit algorithms are particularly good at combining exploration with exploitation and have seen success in many scenarios where this combination is important. This is the case with SARS-Cov-2, where exploration (random sampling) has to be effectively combined with exploitation (contact-tracing, location-based sampling) in a dynamic manner based on data.  One of the key aspects of our work is that we identify specific individuals to test/sample, as opposed to criteria or covariate-based sampling strategies. 

Furthermore, our algorithm operates in a high uncertainty network environment. Specifically, the contacts between individuals are initially only partially observable and are revealed gradually based on individual sampling. The bandit algorithm therefore first uses the Thompson sampling strategy to trade off between expansion in unobserved nodes to identify possible new hot spots, vs. densification on the observed portion to utilize the knowledge learnt for testing. If densification is chosen, an inner level upper confidence bounding (UCB) policy is designed to balance between the individuals having higher probability of infection vs. those having greater information uncertainty. To quantify the closeness of individuals and risk of getting infected, we construct heterogeneous network embedding to encode the social interactions by a continuous latent representation. In order to model how the sampling affects, and in turn is affected by, the underlying environment, we implement an agent-based model constructed from realistic data from New York City.

Several recent Covid-19 studies have focused on the ability to predict and flatten the spread of the disease by applying different intervention policies \cite{Keskinocak2020.07.22.20160036,Keskinocak2020.04.29.20084764, topirceanu2020centralized,atkeson2020will,ferguson2020report,chang2020modelling}. Among the policies that were examined were case isolation, home quarantine, social distancing, restrictions on air travels, and school closures and re-opening. To assess the efficiency of the restrictions, most of these studies coupled S/ I/ R (stands for susceptible / infected / recovered) diffusion modeling \cite{hethcote2000mathematics,cooper2020sir}, with Agent Based simulation Model (ABM) \cite{perez2009agent} of infectious spread under different restrictions scenarios. Among these studies, an ABM of the Covid-19 spread in the city of NY under different quarantine policies was offered by \cite{hoertel2020facing}. This model serves as a basis for the ABM that we developed in this work. Also in contemporaneous work, \cite{grushka2020framework} present a framework that uses historical data to build a classifier that computes a ``risk score'', which is the basis of determining how to combine exploration and exploitation. While the idea of using bandit-based algorithms for testing is important, the challenge is in developing specific methods that address all the complexities noted previously. Our work presents a specific method that is adaptive, dynamic and comprehensive. Along with the lines of whom to sample for testing, there is a similar question of whom to vaccinate. \cite{chen2020allocation}, for example, present adaptive and dynamic covariate-based policies for vaccination. Our framework allows vaccination to be incorporated  into the background dynamics in order to test the effectiveness of combinational strategies such as smart testing with vaccination.

We make the following important contributions. First, we present a novel active sampling algorithm to effectively manage pandemics such as COVID-19 (as far as we know there are no methods in the literature developed for this problem yet). We develop a general multi-armed bandit framework for this problem that (a) can leverage information of different types such as individuals and locations, (b) can handle uncertainty in both the underlying disease dynamic as well as information and (c) is built to sample based on individuals and not covariates. Second, we make several modeling contributions related to how bandit based algorithms can work in this setting. Specifically, we build a heterogeneous network to capture social relations, neighborhood similarity and community membership and show how this can drive a two-level active sampling strategy to sample individuals. The use of network embedding ideas within the heterogeneous network is novel. Third, we make important contributions to the literature on multi-armed bandits as well. Beyond classic bandit problems, there exists limited literature on active search on graph, with the objective of finding as many target nodes as possible with some given property. Most of the existing work assumes that the complete network structure and underlying process is known beforehand. There is very little work on partially observed and dynamically changing networks and underlying (disease) dynamics. We model the partially observed scenarios within this framework and show how the network can be strategically expanded over time to support the active sampling strategy. 

Before presenting the details, it is worth emphasizing that smart testing strategies are needed even if tests are cheap and/or vaccines are available for the following reasons. 
\begin{itemize}
    \item Tests are never truly free, there is an infrastructure behind the testing (generating the tests, distributing, obtaining and storing results) that is expensive.
    \item In active infections that have high economic impact, policy makers may need to provide incentives to make individuals take these tests, particularly if they are asymptomatic (e.g. in the US this may require some compensation for the time an individual who tested positive might have to be in quarantine if the government mandated the test).
    \item For different reasons, some individuals have higher propensity of being infected yet asymptomatic, or pose higher risk on the society when infected. A smart testing strategy can help decision makers detecting these individuals.
    \item Pandemics such as COVID-19 are likely going to recur in the future, with no guarantees that tests for future infections will be cheap (particularly in the early stages, which active testing can play a critical role). The methodology presented in this paper applies to mitigating the spread of any pandemic.
    \item We view vaccines as complementary to testing, and, as we show, active testing strategies remain important even with vaccinations as a combinational strategy to eradicate pandemics. 
\end{itemize}


\section{Problem Formulation}\label{sec:problem}

We approach the problem from the perspective of policy makers who can enact sampling policies. Below we describe a general setting from this perspective that can apply to any disease scenario. COVID-19 specific instantiations of these are presented as brief examples here, but those are discussed in more detail in Sections \ref{sec:ABM} and \ref{sec:MAB}. As we present the setting considered here, we distinguish between information known to the policy-makers and the true state of information in the world. As we will see later in the paper, this distinction enables us to design and run carefully constructed agent-based models to evaluate sampling policies.

Since disease spread occurs in a spatio-temporal manner in a population, the setting we consider has three components. The first component, \emph{data setting}, represents knowledge about the individuals, their spatio-temporal behavior along with characteristics of the physical area. The second component, \emph{process setting}, represents knowledge about the underlying disease and its mechanism of spread. The third component, \emph{policy setting}, represents what policy makers are assumed to have access to, and accordingly what the sampling algorithm will use. These are described below.

\noindent \\ Data setting:
\begin{itemize}
    \setlength\itemsep{0em}
    \item a population of individuals $P$, with set of covariates $Z$ (e.g., age, gender),
    \item a set of locations $L$ in the city,
    \item spatial-temporal data about individuals in $P$ over $L$.
\end{itemize}

\noindent \\ Process setting:
\begin{itemize}
    \setlength\itemsep{0em}
    \item initial actual disease state (e.g., S/E/I/R) of each individual in $p \in P$, 
    \item a spatio-temporal model of contagion that represents how the disease spreads in the population,
\end{itemize}

\noindent \\ The policy makers' setting involves two components, the information setting and the policy setting. The information setting for the policy maker includes:
\begin{itemize}
    \setlength\itemsep{0em}
    \item complete list of individuals $P$, with partial covariates information,
    \item complete list of locations $L$, 
    \item initial disease state of each individual in $P$ (this can be incomplete),
    \item the contagion process is partially known - the policy maker (algorithm) is assumed to be aware of only the high-level factors affecting disease diffusion, such as contacts with exposed individuals or the role of locations where infected individuals have been. As information about a disease is more well-known (such as the specific impact of covariates on disease spread, or the exact probabilities based on locations) we can augment this setting to reflect this information, but don't assume this to be the case in this paper.
\end{itemize}

\noindent The policy setting includes:
\begin{itemize}
    \item a quarantine policy $Q$, that determines when and how individuals will isolate in the population, 
    \item an objective function $O(T)$. This paper focuses on sampling strategies for bridging the gap between the actual infected population and the known infected individuals, as a step towards controlling disease spread. 
    \item constraints on number of tests per day ($D_{max}$).
    \item Sampling an individual to test reveals the following:
        \begin{itemize}
            \item disease state of the individual being tested.
            \item spatio-temporal data about the individual, if infected. In some cases (e.g. access to full GPS data) this will be complete information on the places visited and the actual contact network. If this is by self-reporting (as is common in most places) then this information will be incomplete.
        \end{itemize}
\end{itemize}

Table \ref{tbl:notation} summarized the common notations we use through the paper, and in which section each notation first appears. 

\begin{table}[]
\centering
\begin{tabular}{p{0.15\linewidth} p{0.2\linewidth} p{0.7\linewidth}}
\toprule
\textbf{Section}                         & \textbf{Notation}                                                        & \textbf{Description }                                                                                                             \\
\midrule
Section 2            & $P = \{p_i\}$                                                      & Population                                                                                                               \\
                                 & $Z= \{z_{ip}\}$                                                      & Individuals’ covariates (e.g. age, gender)                                                                               \\
                                 & $L = \{l_j\}$                                                      & City layout (list of locations, e.g. a specific school, restaurant)                                                      \\
                                 & $T = \{t_l\}$                                                      & Time in days                                                                                                             \\
                                 & $Q$                                                               & Quarantine policy (e.g. 14-day isolation for positively tested individuals)                                              \\
                                 & $O(T)$                                                            & High level objective (e.g. minimizing total deaths)                                                                      \\
                                 & $D_{max}$                                                            & Constraints of number of daily tests                                                                                     \\
\midrule
Section 3.1      & $K$                                                               & Number of arms in multi-armed bandit algorithms                                                                          \\
                                 & $x_t$                                                              & Selected arm at time $t$                                                                                                   \\
                                 & $r_t(x_t)$                                                          & Payoff at time $t$ for selecting arm $x_t$                                                                                    \\
\midrule
Section 3.2 & $\mathcal{G} =  <V,E,Z,D>$ & Network of individuals (in the contact network) and locations (in the heterogenous network) connected based on proximity \\
                                 & $V = \{v_i\}$                                                      & Nodes in the graph                                                                                                       \\
                                 & $E = \{e_{ij}\}$                                                     & Edges in the graph                                                                                                       \\
                                 & $D = \{d_i\}$                                                      & Disease states (e.g. S/E/I/R )                                                                                           \\
                                 & $y_v$                                                              & Whether the individual is infected (partial disease state)                                                               \\
                                 & $\mathcal{G}_t$                                   & Network at time $t$                                                                                                        \\
                                 & $t$                                                               & Known network at time $t$                                                                                                  \\
                                 & $f(v)$                                                            & Reward of testing node $v$                                                                                                 \\
                                 & $U/ U_t$                                                           & Subset of nodes to sample from / at time $t$                                                                               \\
                                 & $f$                                                               & Function that maps the node to [0,1] as the probability of infection.                                                \\
                                 & $F(U, f)$                                                         & Set function on subset $U$                                                                                                 \\
                                 & $N$                                                               & Neighborhood in the heterogenous network, a set of nearby locations                                                      \\
                                 & $TYPE_v$                                                           & Set of node types (e.g. individual, location)                                                                            \\
                                 & $f_v$                                                              & Mapping from $v$ to $\{TYPE_v\}$                                                                                              \\
                                 & $TYPE_e$                                                           & Set of edge types (e.g., strong contact, nearby locations)                                                               \\
                                 & $f_e$                                                              & Mapping from $e$ to $\{TYPE_e\}$                                                                                              \\
\midrule
Section 3.3     & $x_v$                                                              & Node embedding                                                                                                           \\
                                 & $DIST$                                                            & Euclidean distance between network embedding of nodes                                                                    \\
                                 & $\sigma_t$                                          & Confidence interval to represent the uncertainty of estimates                                                            \\
                                 & $N_k(v)$                                                           & $k$-neighborhood of node $v$                                                                                                 \\
\midrule
Section 4    & $\tau$                                                              & Simulation epoch in days                                                                                                 \\
                                 & $A = \{a_i\}$                                                      & Agents that represent the population                                                                                     \\
                                 & $AL$                                                              & Represents which location each agent tends to visit and how often     \\                                                  
\bottomrule
\end{tabular}
\caption{Frequently Used Notations}
\label{tbl:notation}
\end{table}

\section{Active Sampling Framework and Algorithm}\label{sec:MAB}

In Section \ref{sec:intro} we identified five unique characteristics of COVID-19. The solution approach presented here builds on ideas from the multi-armed bandit literature and heterogeneous network embedding literature to model these five characteristics. Specifically, the multi-armed bandit framework offers a general approach to combine exploration and exploitation (i.e. sampling neighbors of infected individuals and simultaneously exploring broadly to identify asymptomatic cases as well). Strategies for populating the arms of the bandit framework allow for re-testing individuals who may be repeatedly exposed as well, addressing another key requirement. The need to sample individuals specifically (as opposed to covariates) and leveraging locations is addressed in our work through a heterogeneous network embedding framework, which is used by the sampling algorithm. We present these details as follows in this section. First, we present a high-level overview of the multi-armed bandit framework and related literature. We then present the hetergeneous network embedding framework that is used to model individuals and locations. Finally, we present the sampling algorithm used to dynamically identify specific individuals to test.

\subsection{Background and Literature Review}
The Multi-Armed Bandit (MAB) is a generic framework to address the problem of decision making under uncertainty \cite{auer2002finite,langford2008epoch}. In this setting, the learner must choose from among a variety of actions and only observes partial feedback from the environment, without prior knowledge of which action is the best. In the classic stochastic K-armed bandit problem,  at each time step $t$, the learner selects a single action/arm $x_t$ among a set of $K$ actions and observes some payoff $r_t(x_t)$.  The reward of each arm is assumed to be drawn stochastically from some unknown probability distribution. The goal of the learner is maximize the cumulative payoff obtained in a sequence of $n$ allocations over time, or equivalently minimize the \emph{regret} \cite{bubeck2012regret}, which is defined as the difference between the cumulative reward obtained by always playing the optimal arm  and the cumulative reward achieved by the learning policy,
$$\mathcal{R}_n = \max_{i = 1,...,K}\mathbb{E}\big{[}\sum_{t = 1}^n r_t(i) - \sum_{t = 1}^n r_t(x_t)\big{]}.$$

The fundamental exploration/exploitation dilemma is to: (1) gain as many rewards as possible in the current round, but also (2) have a high probability of correctly identifying the better arm. Many MAB algorithms with near-optimal guarantees have been proposed and applied in various domains. Out of all existing approaches to MAB, Upper Confidence Bound (UCB) policies \cite{auer2002finite, garivier2011kl} are the most popular approaches that are designed based on the principle of ``optimism in the face of uncertainty'', with arms chosen based on an upper confidence bound on the expected reward for each arm. Other approaches include Thompson sampling \cite{chapelle2011empirical, thompson1933likelihood}, expected improvement \cite{huang2006global,picheny2013benchmark} and knowledge gradient policies \cite{frazier2008knowledge}.

Beyond classic bandit problems, there exists literature on active search, with the objective of finding as many target nodes as possible with some given property. Most of the existing work assumes that the complete network structure is known before hand. For example,  \cite{ma2015active} models the fully observed network using Gaussian random fields  and proposes a UCB-type policy based on sigma optimality. Limited attention was paid to partially observed networks. \cite{bnaya2013social} proposed modeling the problem of social network querying and targeted crawling as an MAB problem with the goal of adaptively choosing the next profile to explore. \cite{singla2015information} deals with the case where each node’s visibility is limited to its local neighborhood and new nodes become visible and available for selection only once one of their neighbors
has been chosen. However, it is assumed that exploring a node reveals its 2-hop neighborhood which is usually not feasible in real-world social networks. \cite{soundarajan2017varepsilon} proposes $\epsilon-$WGX to solve the Active Edge Probing problem in incomplete networks, and yet in their case, a node can be queried multiple times and a single random edge adjacent to the queried
node is revealed  in each query. \cite{madhawa2019multi} proposes a $k$NN-UCB policy for partially observed network based on structure graph features including degree, average neighbor degree, median neighbor
degree and average fraction of probed neighbors.

In the MAB context of active search on graphs, the question our research addresses can be posed, more generally, as:
{\it given a partially observed network with no information about
how it was observed, and a budget to query the partially observed network nodes, can we
learn to sequentially ask optimal queries?} This is a novel extension to the MAB literature.


\subsection{The Sampling Framework}\label{network_embedding}

COVID-19 is a pandemic that spreads via social contacts, directly or indirectly through locations. There are two approaches that can be taken to model this. The first approach, which we present below, is based on using contact networks only. This is close to how contacts are traced today and might therefore be immediately applicable. The second approach, which we present further below, leverages locations to build \emph{heterogeneous} networks that combine individuals and locations. This is useful when both contact information as well as information about locations are available.

Contact networks are modeled as undirected, time dependent, network $\mathcal{G}_t = <V, E, Z, D> $, 
where $V$ is the node representation of the individuals $P$,  with covariates information $Z$. The network is evolving over time with new contacts reported to the policy maker. For example, if the policy maker aims to make daily decisions, $\mathcal{G}_t$ represents the daily snapshot of the network on day $t$.
An edge $e_{ij} \in E$ represents a direct contact between individuals $v_i$ and $v_j$. At any time point, every node $v$ in our graph $\mathcal{G}_t $ has one disease state in $D$. We use $y_v$ to represent whether the individual is tested positive. The actual disease state can never be fully observed (e.g. the difference between susceptible,  exposed and recovered cannot be identified).

The graph $\mathcal{G}_t$ is not fully observable and only partial information $\hat{\mathcal{G}}_t $ is available to the decision maker. When a node tests "positive", its \textit{known} contacts will be revealed at once, with the understanding that there might be other contacts that are latent and not readily observed.

Mathematically, this problem of active sampling can be formalized as a stochastic combinatorial optimization problem with $<V, \mathcal{U}, f>$, where $V$ is the set of nodes to be sampled from, $\mathcal{U} = \{U\subset V: |U| \leq D_{max} \}$ is a family of subsets of $V$ with up to $D_{max}$ (testing capacity) nodes, and $f$ a function that maps the node to [0,1] as the probability of infection. The objective function for all the nodes in a set $U \subset V$  is defined by $F(U, f)$. For example, if the objective is to find as many infected nodes as possible in each day, $F(U, f)$ can be defined as:
$$F(U,f) = \sum_{v \in U}f(v).$$ 
The goal of the sampling policy is to adaptively select a sequence of subsets $U_t$ to test, in a way that maximizes the cumulative rewards of $F(U_t,f)$ over time, recognizing that we observe, on testing, the realized reward $f(v)$ of each node $v \in U_t$ immediately, i.e. whether the tested individual is infected or not ($y_v$).



An important observation is that immediate contacts alone may be insufficient to consider for COVID-19 testing and we need a richer representation of similarity between individuals. We seek to model \emph{similarity} such that if one agent is infected, then a \emph{similar} node is also possibly infected. This can help identify potential regions at high risk of outbreaks given previous outbreak locations, the mobility of agents, and their network structure. Local structure features, like degree, number of triads, and centrality, have been used in previous network analysis on finding structurally similar nodes \cite{Henderson}. Although these features can help infer roles of each node, such as super spreaders, or periphery nodes, they do not capture the information about the neighborhood similarity, social relations, and community membership. For example, if two agents are living in the same household,  COVID-19  can easily spread through contact  transmission or droplet transmission. Due to large amount of pre-symptomatic and asymptomatic disease transmissions,  even if there is no direct contact link between two agents, if they constantly went to the same supermarket, or live in nearby neighborhoods, the agents can end up infecting each other. This complex interplay of social relationships and locations has resulted in several infections from the same gathering (super-spreaders, hotspots) and brought up the importance of even seemingly minor occurrences such as how even a few agents with the same travel history may have caused unknown breakouts along the way.

Our goal is to quantify these above mentioned social relations by a continuous latent representation of nodes, which can then be exploited to guide the sampling policy to more effectively allocate testing kits of limited capacity. A prioneering method, DeepWalk \cite{perozzi2014deepwalk}, uses language modeling approaches to learn latent node embedding in the following two steps (please refer to Appendix A for the technical details of this approach). First it traverses the network with random walks to infer local structures by neighborhood relations, and then uses a SkipGram model \cite{mikolov2013distributed} to learn node embedding based on the produced samples. In our context, by translating the nodes in the network into a continuous space this way, the node embedding thus generated provides a method to efficiently sample individuals to test based on how ``close" they are to others.

The above discussion is restricted for representation learning for contact networks where nodes and relationships are of a singular type. However, the network embedding ideas help in ``indirectly" modeling locations (e.g. ``similar" individuals constructed as described above could have been similar because they went to the same location). Still, if location information was available to the policy maker then explicitly modeling is useful. Specifically, it is well-known that in the case of COVID-19, the transmission patterns and severity are largely depending on the type of the contacts. For example, nursing homes and long-term care facilities are especially vulnerable to the prevalence and spread of COVID-19  because they combine numerous risk factors for transmission: elderly people with underlying health conditions, congregate living, short of medical resources, frequent staff and visitors entering and leaving facilities, inadequate staffing,  and infection control for an emergency. While contact networks capture whether two individuals are in close proximity in a certain period of time, it does not include the the type of the contact and/or the locations where the individuals met.  In order to be able to identify the specific locations and/or gathering of the COVID breakouts, we augment the contact network with diversity of node types  and multiple types of relationships between nodes.  

\begin{figure}[htp!]
 \centering
  \includegraphics[trim=150 140 30 200, clip, width=\textwidth]{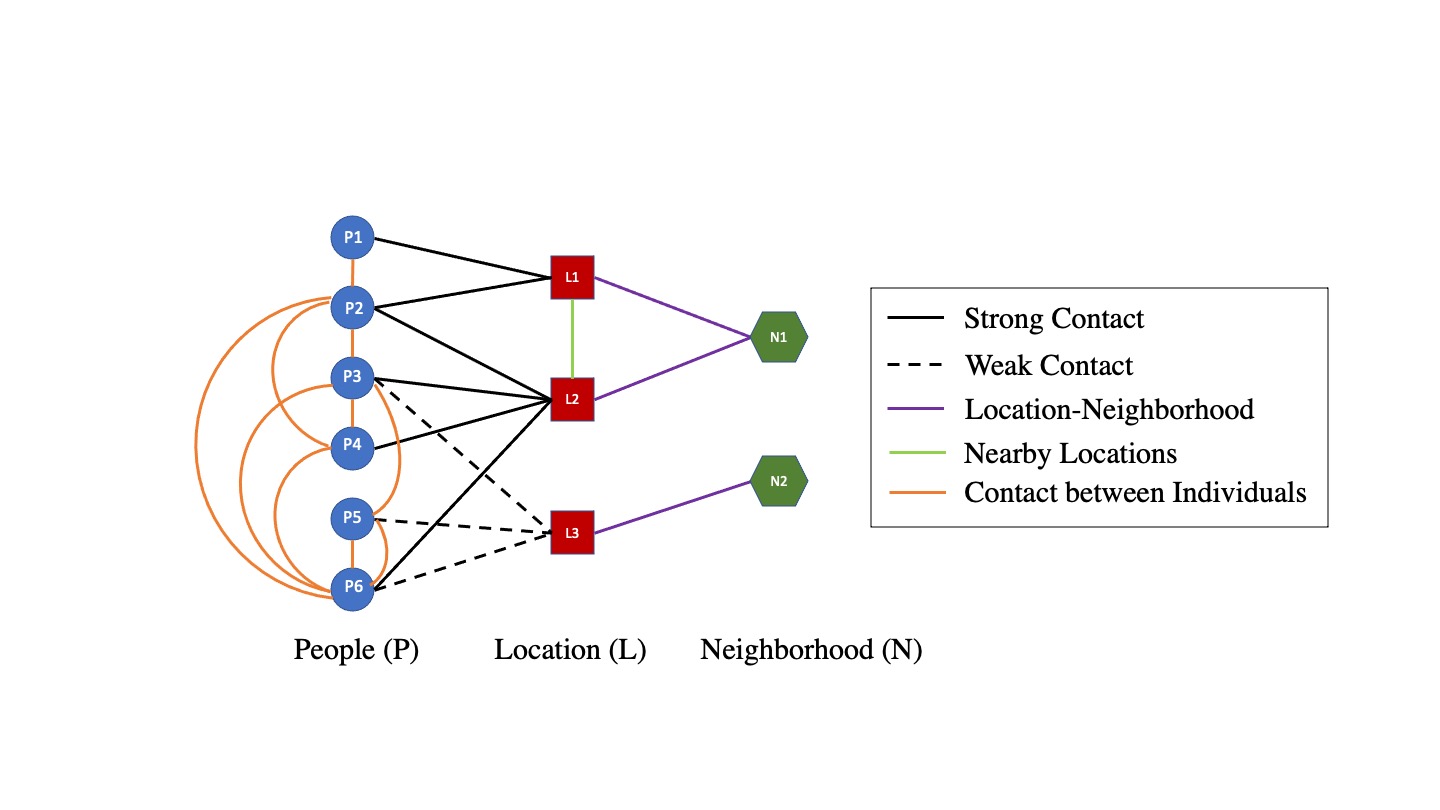}
 \caption{An illustrative example of a heterogeneous contact network.}
 \label{fig:network}
 \end{figure}

Specifically, as illustrated in Fig. \ref{fig:network}, we represent the contact information with people (P), locations (L), neighborhood (N) as nodes. We also augment this network with edge types that reflects the number (or frequency) of contacts, or the intensity of each dyadic interaction. For example, individuals from the same household will have a much stronger interaction than the contact between random people that potentially met at the same subway station. The link between locations and neighborhoods is intended to generalize the observation of the transmissions to local neighborhood, even if two individuals are not reported in contact tracing due to unawareness. 
\begin{definition}[heterogeneous network]
A heterogeneous network is a network $\mathcal{G}$ with multiple types of nodes or multiple types of edges. Formally, each node $v$ and each edge $e$ are associated with a type mapping function $f_v: v \rightarrow \text{TYPE}_v$, and $f_e: e \rightarrow \text{TYPE}_e$, where $\text{TYPE}_v$ and $\text{TYPE}_e$ are the set of node and edge types, respectively.
\end{definition}

Heterogeneous networks present unique challenges that cannot be handled by node embedding models that are specifically designed for homogeneous network, such as the DeepWalk method mentioned above \cite{dong2017metapath2vec, chang2015heterogeneous}. When performing the random walk, DeepWalk ignores their node types. However, it is demonstrated that such random walks on heterogeneous networks are biased to highly visible types of nodes and concentrated nodes.  With this regards, Metapath2vec \cite{dong2017metapath2vec} extends DeepWalk to heterogeneous networks by incorporating meta-paths which have been shown to be effective in many data mining tasks in heterogeneous information networks \cite{sun2013pathselclus,dong2015coupledlp, fu2017hin2vec}. 

A meta-path is a sequence of node types encoding key composite relations among
the involved node types. Different meta-paths express different
semantic meaning, and thus different disease transmission pathways.  For example, in Fig \ref{fig:network}, a meta-path "P-L-P" represents the direct interactions of two  individuals in the same place, e.g. household, workplace, school. A meta-path "P-L-N-L-P" represents two individuals residing in the same neighborhood. Then the random walk is confined on the pre-defined meta-paths. Additionally, we extend the metapath2vec by assigning different weights to edges corresponding to the contact intensity, which in turn translates to the transition probability in the biased random walk. For example, in a homogeneous network and traditional random walk, the next node of $P_3$ can be all types of nodes connected to it - $P_2, P_4, P_5, L_6, L_2, L_3$. However, under the meta-path "P-L-P", the random walk can only result in location nodes (L) - $L_2, L_3$, given that itself is an individual node (I). Meanwhile, the transition probability from $P_3$ to $L_2$ is higher since their link indicates a more intensive relationship. The meta-paths commonly requires that the starting nodes and ending nodes are of the same node type so that random walk can be performed recursively.

After the random walk sequences are generated, a modified heterogeneous SkipGram is used considering node type information during model optimization.


\subsection{The Sampling Algorithm}
In the classic stochastic multi-arm bandit problem, the learner selects one of the $K$ arms at each time step and receives a stochastic reward sampled from some unknown reward distribution. In our case, each node is one possible arm. Due to finite but large number of arms, feature-based exploration is needed to share the knowledge learnt for similar nodes.  In this paper, we formulate the active sampling problem as a contextual bandit, in which before making the choice of action,  the learner observes the currently known disease states, and a feature vector $x_i$ associated with each of the possible arm (a.k.a. individual node $v_i$) encoded from the contact network structure. The learner then selects an action and reveals whether the node is a search target. 

As suggested by \cite{chen2020allocation, long2018spatial}, it is challenging to compute the optimal dynamic allocation using dynamic programming even for 35 age-compartment pairs and the heuristic obtained from approximate dynamic programming performs worse than simple heuristics such as single-step myopic policy.  Hence full learning model at individual level is challenging, if not infeasible. In the same time, there are many unobserved positive cases  affecting the disease transmission, as well as un-reported (random) contacts. With this regards, we treat the graph context drawn $i.i.d.$ from some context density.

It combines two major challenges. First, the number of possible actions grows exponentially with the cardinality constraint $D_{max}$. Second, the policy maker can only observe the partially reported portion of the network. 

To deal with the partially observed network, we thus consider two levels of explorations:
\begin{itemize}
\item Given the current observed network $\hat{\mathcal{G}}_t$, the probability of infection is not known beforehand, and needs to be learnt as we gradually sample individuals and observe their test results.
\item Expansion on (randomly selected) unobserved nodes to identify new hotspot.
\end{itemize}
Hence the proposed sampling policy concerns two levels of exploration/exploitation tradeoff. 

\paragraph{Outer Level: expansion vs. densification.} The outer level is modeled as a two-armed bandit, corresponding to the two choices of expansion and/ or densification. Consistent with the literature, we choose to use the Thompson sampling policy with Beta-Bernoulli distribution \cite{thompson1933likelihood, chapelle2011empirical}.  The expected reward $\theta$ of the two choices are modeled with a Beta distribution. After one choice is chosen, the realization of the reward (whether an individual is tested positive) is sampled from a Bernoulli distribution.  With a testing capacity constraint $D_{max}$, we repeatedly sample $D_{max}$ realizations of $\hat{\theta}$ and choose between the two choices of expansion and/or densification that has the highest sampled value. If expansion is chosen, our policy samples a node from the network that was not in $\hat{\mathcal{G}}_t$ uniformly at random. If densification is chosen, our policy use the inner level exploration/ exploitation algorithm to choose individuals to sample. We use $\bar{D}_{max}$ to represent the total number of tests that are allocated to densification. After the test result $y_v$ for each chosen individual is revealed, the posterior Beta distribution will be updated. 

\paragraph{Inner Level: individual selection under combinatorial optimization: combinatorial $k$NN-UCB.}
On each day, we first use network embedding $x_v$ from Section \ref{network_embedding} to find the node representation based on the past 14-days contacts of each known individual. The node embedding encodes the  network structure, i.e. neighbors and connection information, and quantifies the similarity between two nodes in the sense that if one is infected then the other is highly likely to be infected from potential contact chains. Our goal is to effectively allocate $\bar{D}_{max}$ lab tests to individuals to alleviate the disease spread. 

The goal of the inner level bandit algorithm is to adaptively select a sequence of subsets $U_t \subset V$ to test and maximize the cumulative rewards of $F(U_t,f)$ over time. We also observe the realized reward $f(v)$ of each node $v \in U_t$, i.e. whether the tested individual is infected or not.

In this paper, to deal with the challenge that the  number  of  possible  individuals to sample is a combinatorial set given the  the cardinality constraint $\bar{D}_{max}$,
 we allow the policy maker to use any exact/ approximation/ randomized algorithm, termed as ORACLE, to find solutions for 
$$U^{\text{OPT}} \in \arg \max_{U \in \mathcal{U}} F(U, f) = \arg \max_{U \in \mathcal{U}} \sum_{v \in U}f(v).$$ We denote the solution as $U^* = \text{ORACLE}(V, \mathcal{U}, f, \bar{D}_{max})$. 

Given that the contextual information is presented in the format of a network rather than numeric feature vectors, we thus adopt non-parametric strategies and use $K$NN-UCB for structured bandits. Specifically, we define the $k-$nearest neighbor upper confidence bound as:
\begin{equation}\label{ucb}
\hat{f}(v_i) + \eta\sigma_t(v_i),
\end{equation}
where the expected reward of node $i$ is estimated with weighted $k$NN regression as
 $$\hat{f}(v_i) = \frac{1}{k}\sum_{v_j \in \mathcal{N}_k(v_i)} \frac{y_j}{\text{DIST}(x_i, x_j)},$$ 
 and $\text{DIST}(x_i, x_j)$ is the euclidean distance between network embedding of node $i$ and $j$, and $\mathcal{N}_k(v)$ is the $k$-neighborhood of node $v$. The uncertainty is chosen to be the average distance to points in the $k$-neighborhood,
$$\sigma_t(v_i) = \frac{1}{k}\sum_{v_j \in \mathcal{N}_k(v_i)}\text{DIST}(x_i, x_j).$$

If at each time step, only one arm is selected, the traditional UCB policy selects the arm with the highest upper confidence bound index, i.e. Eq. (\ref{ucb}). However, in the case of combinatorial optimization, we propose the algorithm Combinatorial $k$NN-UCB which makes use of the $\text{ORACLE}(V, \mathcal{U}, f, \bar{D}_{max})$ to provide (approximate) solutions for the offline optimization problem. The pseudocode is  provided in Algorithm \ref{comKNN}. Due to the fact that the number of individuals is huge,  scalable space partition method is needed to restrict the search of nearest neighbors to local partition.

 \begin{algorithm}[htp!]\label{comKNN}
\caption{Pseudo-code for Active Sampling}
\SetAlgoLined
\SetKwInOut{Input}{input}\SetKwInOut{Output}{output}

 \Input{Past 14-day heterogeneous network $\hat{\mathcal{G}}_0 = <V, E,Z, D>$,  oracle ORACLE, trade-off parameter $\eta$, daily constraint $D_{max}$, reward function $f$, $\alpha$, $\beta$ for prior Beta distribution, $S_i = 0$, $F_i = 0$  for $i = 1,2$ }
 \For{$t=1$ to $T$}{
 Update the contact network $\hat{\mathcal{G}}_t$ and find the network embedding of each node $x_v$\\
 initialize the proportion allocated to densification $\bar{D}_{max} = 0$\\
 \For {$l == 1$ to $D_{max}$}{
 \textbf{[Outer Level]}\\
Draw  $\hat{\theta}_i$ from $Beta(S_i + \alpha, F_i + \beta)$ for $ i = 1,2$\\ 
Choose arm $a_t=\arg\max_{i}\hat{\theta}_i$\\
\If{$a_t$ == expansion}{Randomly sample a node $v$ that is not in $\hat{\mathcal{G}}_t$ to test, and update $y_v$}
\Else{$\bar{D}_{max} = \bar{D}_{max} + 1$}
}
\textbf{[Inner Level]}\\
 Compute the UCB for each individual (P) nodes in $\mathcal{G}_t$ as
 \begin{eqnarray*}
 \hat{f}(v_i) &=& \frac{1}{k}\sum_{v_j \in \mathcal{N}_k(v_i)} \frac{y_j}{\text{DIST}(a_i, a_j)}\\
 \sigma_t(v_i) &=& \frac{1}{k}\sum_{v_j \in \mathcal{N}_k(v_i)}\text{DIST}(a_i, a_j)\\
 \bar{f}_t(v_i) &=& \hat{f}(v_i) + \eta \sigma_t(v_i)
  \end{eqnarray*}\\
  Compute $U_t \leftarrow \text{ORACLE}(V, \mathcal{U}, \bar{f}_t, \bar{D}_{max})$\\
  Test individuals in set $U_t$\\

  Observe the testing results and update $y_v$\\
  Compute the number of positive cases under expansion and densification, respectively, $\bar{S}_1, \bar{S}_2$\\
  Compute the number of negative cases under expansion and densification, respectively, $\bar{F}_1, \bar{F}_2$\\
Update posterior Beta distribution:
\begin{eqnarray*}
S_i &=& S_i +\bar{S}_i\\
F_i &=& F_i +\bar{F}_i, \text{ for } i = 1,2
\end{eqnarray*}

}
\end{algorithm}

\section{Agent-Based Model}\label{sec:ABM}

In the real-world setting, as long as we have daily spatio-temporal data on individuals in a population, the algorithm presented in the previous section can be used to select individuals to test. However, and somewhat counter-intuitive, real-world data alone is insufficient to fully understand the impact of different sampling policies. The "real world" represents a single run of how reality is shaped, and does not offer the advantage of running experiments to test counterfactuals. For instance, would the sampling algorithm have worked as well under a different quarantine policy, or under various compliance scenarios? Agent-based models, on the other hand, are particularly effective to answer these types of more general questions.

In this paper we couple a disease progression model \cite{hethcote2000mathematics} with Agent Based simulation Model (ABM) \cite{perez2009agent} of infectious spread under different settings. Disease progression is commonly modelled with the deterministic compartmental S/I/R \cite{hethcote2000mathematics,cooper2020sir}, which divides the population into three compartments - susceptible to the disease (S), actively infected with the disease (I), and recovered (or dead) and no longer contagious (R)) - and defines transmission rates between the compartments.  To account for additional aspects of the Covid-19 transmission, we use an extended S/I/R model that includes individuals that are exposed (E) to the disease \cite{silva2020covid,prem2020effect}, and asymptomatic (A) individuals \cite{manchein2020strong,koo2020interventions} - an S/E/A/I/R model.

We constructed a data driven stochastic Agent-Based Model (ABM) of the COVID-19 epidemic in New York city. Based on \cite{hoertel2020facing}, our ABM model includes four components: (1) synthetic population with demographic characteristics and spatial information that is generated to resemble the city of New York, (2) daily interaction network between individuals (agents) in the population, (3) disease dynamics that spreads via interactions, and progresses as an S/E/A/I/R model, (4) policy makers, as agents, who influence the environment based on the sampling strategies and quarantine policies in place.

In the reminder of this section we describe components (2) - (4) of the ABM. The population generating approach in component (1) is detailed in Appendix B. In algorithm \ref{alg:ABM} we provide the pseudo code of our model.

\begin{algorithm}
    \caption{Pseudo-code for the ABM}
    \label{alg:ABM}
    \DontPrintSemicolon
    \SetAlgoLined
    \textbf{Input:} \;
    \Indp Agents $A = \{a_1, ..., a_n\}$; Locations $L = \{l_1, .. l_m\}$;
    Agents-locations interaction $AL = \big\{\{\phi_{a_i, l_l}\}\big\}$; 
    Number of initial infected agents $\tilde{n}$; \;
    \Indm \textbf{Output:} \;
    \Indp Disease state $D(A) = \{d(a_1), ... d(a_n)\}$; meetings log; \;
    \Indm \textbf{Initialize:} \;
    (1) Set initial disease state $D(A)$ such that: \;
        \Indp for $\tilde{n}$ agents $d(a) \leftarrow Ia$, \;
        for $N-\tilde{n}$ agents $k(a) \leftarrow S$; \;
    \Indm (2) Set simulation day $\tau \leftarrow 1$; \;
    (3) Initialize set of agents in quarantine $Q \leftarrow \{\}$; \;
    (4) Initialize agent-location log; \;

    \While{some agents are still infected}{
        \textit{// Draw locations per agents:} \;
        \For {$a_i \in A \cap Q$}{
            Draw daily locations $l_{a_i}$ according to the agent-location interactions propensity $AL$; \;
            Append tuples $\big\{\{a_i, l_j \in l_{a_i}\}\big\}, \tau$ to the agent-location log;
        }
        \textit{// Get new infected agents:} \;
        \For {$l_j \in L$}{
            Get the list of agents $a_{l_j}$ that visited $l_j$ at time $tau$ from the agent-location log; \;
            If one or more agents is infected ($d(a \in a_{l_j}) \in \{Ia, Is, Ic\}$, set the disease status of all \textbf{\textit{Susceptible (S)}} agents $\in a_{l_j}$ as \textbf{\textit{Infected asymptomatic (Ia)}};
        }
        \textit{// Progress disease state:} \;
        Progress disease state $d_i$ for all agents according to Figure \ref{fig:MC}; \;
        \textit{// Put critically infected in self-quarantine:} \;
        \For {$d_i \in D$}{
            \Switch{$d_i$}{
                \Case{$Is$: }{Append $a_i$ to $Q$} \;
                \Case{$R$ or $D$: }{
                    Remove $a_i$ from $Q$} \;
            }
        }
        \textit{// Put agents in enforced-quarantine:} \;
        \SetKwBlock{SMPL}{Interact with policy-makers (sampling policy)}{end}
        \SMPL{
            \textbf{Send:} $\tilde{A} \leftarrow $ set of externally sampled agents to be tested\;
            \textbf{Receive:} $\doubletilde{\textit{A}} \leftarrow $ set of agents to be set in quarantine\;
            \textbf{Respond:}  Append $\doubletilde{\textit{A}}$ to $Q$ \;
        }
        $\tau \leftarrow \tau + 1$\textit{// Advance simulation time:} \;
    }
\end{algorithm}

\subsection{Interaction data structure}
Our data is composed of a set of individual agents $A = \{a_1, ..., a_n\}$, a set of geographic locations $L = \{l_1, .. l_m\}$, and the propensity of the interactions between the agents and the locations: $AL = \big\{\{\phi_{a_i, l_l}\}\big\}$ (how often an agent goes to the location). 

Agents in our data is characterized by age, gender, and a list of locations that the agents goes to, at a given probability. Locations can take different location types, and are spread across the neighborhoods of New York City. For this paper we set the following location types: household, workplace, school, station, and supermarket. Two agents may meet if they go to the same location at the same time. The meeting probability at a given location is a function of the location type (for example, agents that live at the same house will meet with a probability of $\phi_{household}=1$, while agents that go the the same supermarket, will meet at a much lower probability: $\phi_{supermarket} << 1$). In addition to the different location types, we add a special location type that we call a "mixing location", in which agents can meet at random. Mixing locations in our model simulates meeting at public spaces, such at parks, shops, theater, etc. 

Note that our sampling algorithm works as an agent within this framework and can be used with any ABM setting - all that is required is for the ABM to simulate contacts and movements of individuals in a population. The above are used primarily for exposition to simulate one specific city (New York). 

Agents in our model can potentially be in one of seven disease states, as described below. We formally define $D(A) = \{d(a_1), ... d(a_n)\}$ to be the set of disease states of all agents.

\subsection{Disease dynamic}
\begin{figure}[h!]
 \centering
  \includegraphics[width=0.9\textwidth]{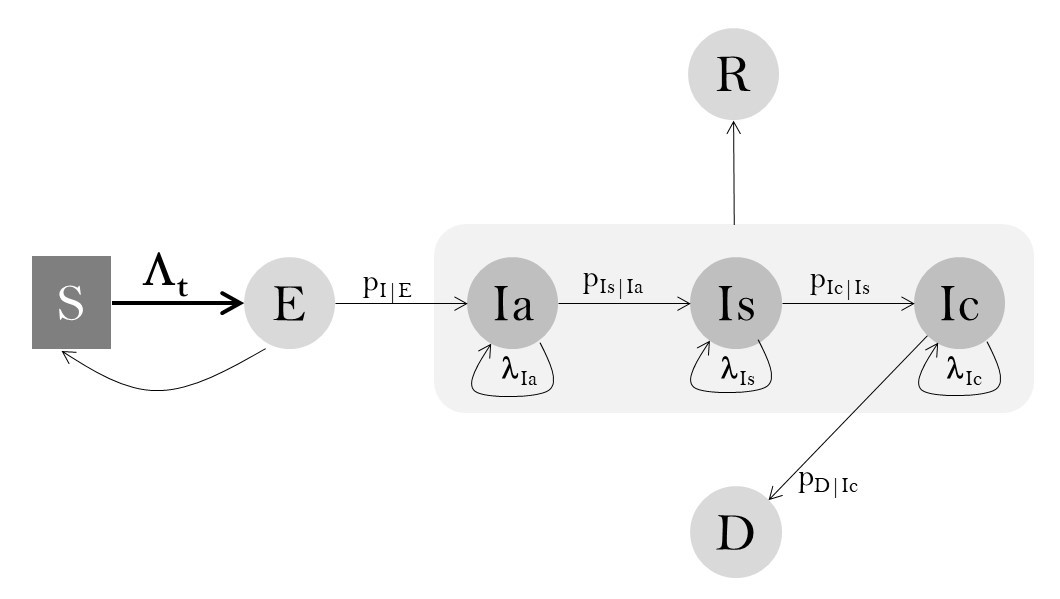}
 \caption{Transmission dynamics of the Corona virus}
 \label{fig:MC}
 \end{figure}

The disease process used in this paper follows the deterministic \textit{S/E/A/I/R} (Susceptible ($S$) - Exposed ($E$) - Asymptomatic ($A$) - Infectious ($I$) - Recovered ($R$)) compartmental model, with death ($D$) rate. We further split the infectious state into three (instead of two) disease stages: Infected asymptomatic ($Ia$), Infected symptomatic ($Is$), and Infected critical ($Ic$). Beyond being a more accurate description of COVID-19, this allows us to flexibly define different policies and different behaviours to infected individuals according to the state of their disease. Similar to \textit{S/E/A/I/R}, we assume Recovered ($R$) is an absorbing state, implying that infected individuals can either die or become immune. Figure \ref{fig:MC} presents the disease states and the transmission dynamics in this paper. 

Transmission probabilities are assumed to be constant across individuals. This is due to the fact that the impact of patient's covariates on the disease dynamics is still unknown. The transmission rate between states $S$ and $E$ is $\Lambda_t$ that determines the rate at which uninfected individuals are exposed to infected individuals. Once exposed, they become infected with probability of $p_{I|E}$. Otherwise they remain susceptible with the complementary probability ($1-p_{I|E}$).
In infected individuals, the disease may escalate and become mild ($Is$), at a probability of $p_{Is|Ia}$, then critical ($Ic$), at a probability of $p_{Ic|Is}$. At any stage of the disease, infected patient can recover. Recovery takes on average $\lambda_{Ia}$ for asymptomatic agents, $\lambda_{Ia} + \lambda_{Is}$ for symptomatic agents, and $\lambda_{Ia} + \lambda_{Is}$ for critically ill agents. The latter disease state can result in death, at a rate of $p_{D|Ic}$.

We note that the sampling algorithm presented in this paper only uses data observed as a result of the actual disease diffusion process, and does not depend specifically on the details of how this is implemented in the ABM. In that sense, our approach will work with any extensions of the model implemented in here.

\subsection{Policy makers as agents}

At the end of each ABM simulation epoch we sample a set of agents to be tested for COVID-19. Once an agent is sampled, her disease state as well as (possibly incomplete) contact network are revealed to policy makers, whom can then decide which agents should be put into quarantine. 
Other than sampled agents, we assume that critically infected agents self quarantine themselves. The quarantine duration is set for 14 days, following the current global practice. While in quarantine, an agent's disease dynamics continues, but her interactions with others stop.



\section{Implications for Policy}

The paper presented a smart-testing approach that was implemented on simulated data. In practice, our method can be used by policy makers to determine whom to test if the following information can be provided: (1) a list of all individuals in the population; (2) a partial mapping of individuals to locations that they can be associated with (e.g., household, workplaces, schools); (3) the set of known infected individuals in the population; and (4) daily-contact network for individuals - ideally, this should be provided for all individuals that were tested positive, but the method can work even with partial information. This daily contact network, for instance, can be computed easily from readily available mobile tracking data in some cases.

Our method can also be used by policy makers to examine carefully constructed scenarios in a world with smart-testing strategies. These scenarios may include different quarantine policies, intervention strategies or social behavior such as compliance or willingness to be vaccinated. The scenarios constructed can also experimentally then show the outcomes under smart-testing versus other testing strategies.

\bibliographystyle{abbrv}
\bibliography{bib}

\appendix
\section{APPENDIX: DeepWalk}

The below are the technical details of DeepWalk \cite{perozzi2014deepwalk} - the use of language modeling approaches to learn latent node embedding in two steps: random walk and word embedding. The analogy made by DeepWalk is that nodes in a network can be thought of as words in a natural language.

\begin{figure}[htp!]
 \centering
  \includegraphics[trim=0 250 600 80, clip,width=0.55\textwidth]{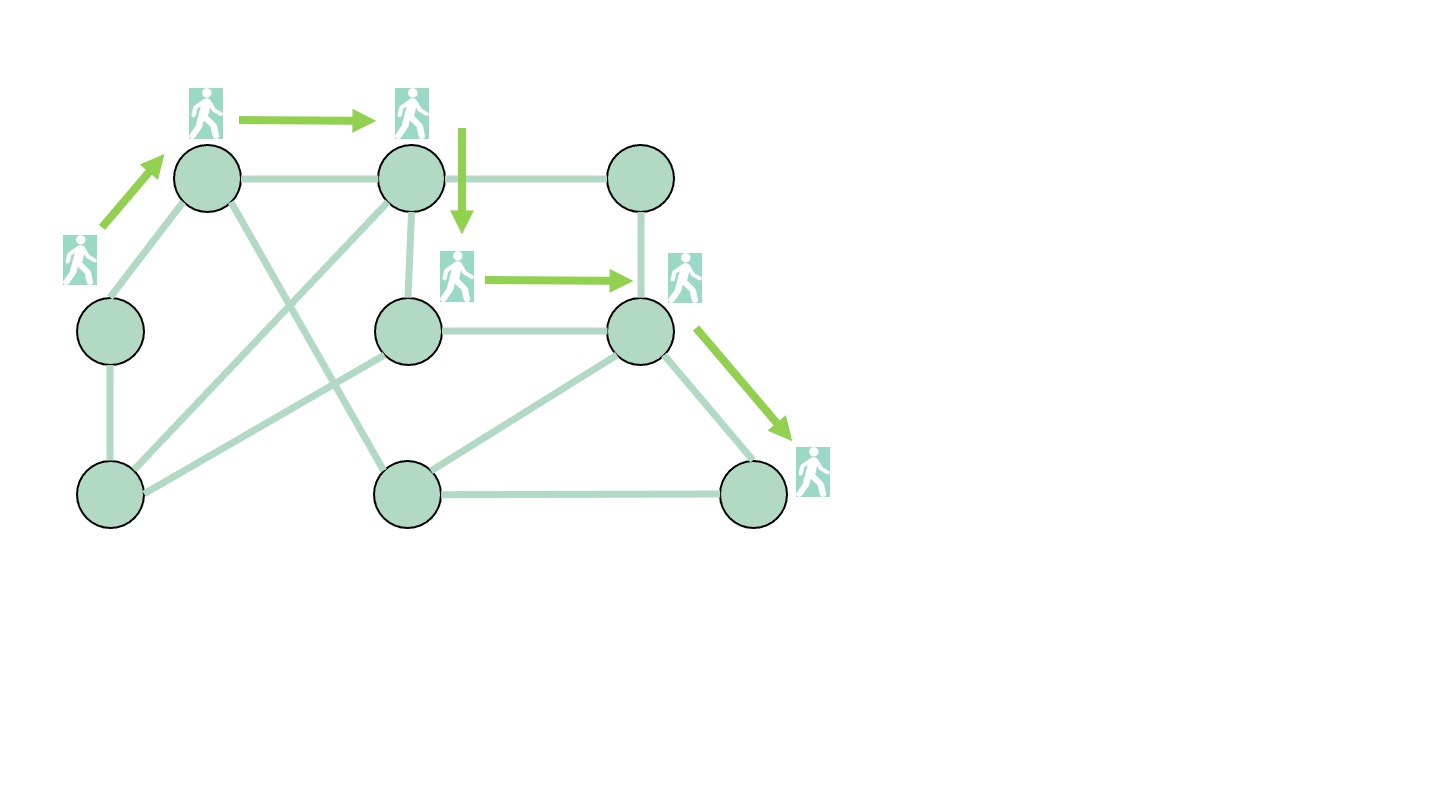}
 \caption{An example random walk of length 6 on a network.}
 \label{fig:random}
 \end{figure}

\paragraph{Random Walk on Graph.}{The first step of DeepWalk is to identify the context nodes for each node. Start at any given node, it identifies  all its neighbors, randomly select one, and walk, with pre-defined path length, as illustrated in Fig. \ref{fig:random}. Repeat until there are enough samples. By generating truncated random walks from the network, the context nodes of $v \in V$ can be defined as the $k$ neighboring nodes in each random walk sequence, which is a combination of nodes from $v$'s 1-hop, 2-hop,..., k-hop neighbors. The idea is that the techniques used to model natural language (where the symbol frequency follows a power law distribution) can be re-purposed for nodes appearing in short random walks whose frequencies also follow a power-law distribution. }

\paragraph{SkipGram.}{SkipGram algorithm is  used to learn word embeddings which represent words in vector space so that if the word embeddings are close to one another,  those words are semantically similar to one other \cite{mikolov2013distributed}. The key idea of SkipGram is to quantify the similarity between any two words by how frequently they share the same surrounding words.  Skip-gram first identifies the context (neighboring) words for a given target word within a pre-defined window size. Then it predicts the context work by maximizing the conditional probability of observing the context words given the target word.

 }

\section{APPENDIX: Data Structure of ABM}
In this section we describe how we generated the synthetic population for the ABM. The synthetic data was generated to describe the city of New York. Specifically, we focus on the Manhattan area, in which the population is large and dense. 

\subsection{Individual characteristics}
Agents in our model were characterized by age and gender, both follow the age and gender distributions in NYC \cite{USCensus}, as depicted in Table \ref{tbl:IndivChar}. These characteristics were later used to define social interactions according to household structures in NYC, age groups, etc.

The number of agents varied in our simulation runs between (1) the entire Manhattan population, which is approximately 2M, (2) a large subset of the population: $\sim$ 100K, (3) a small subset of the population: $\sim$ 20k.

\begin{table}[h]
\centering
\begin{tabular}{p{0.15\textwidth}|p{0.2\textwidth}|p{0.2\textwidth}|p{0.2\textwidth}l}
\cline{2-3}
                                                       & \textbf{Level}     & \textbf{Percentage} &  &  \\ \cline{1-3}
\multicolumn{1}{|l|}{\multirow{13}{*}{\textbf{Age}}}   & \textless{}5     & 6.8\%          &  &  \\ \cline{2-3}
\multicolumn{1}{|l|}{}                                 & 5-9              & 7.0\%          &  &  \\ \cline{2-3}
\multicolumn{1}{|l|}{}                                 & 10-14            & 6.6\%          &  &  \\ \cline{2-3}
\multicolumn{1}{|l|}{}                                 & 15-19            & 6.5\%          &  &  \\ \cline{2-3}
\multicolumn{1}{|l|}{}                                 & 20-24            & 7.4\%          &  &  \\ \cline{2-3}
\multicolumn{1}{|l|}{}                                 & 25-34            & 17.1\%         &  &  \\ \cline{2-3}
\multicolumn{1}{|l|}{}                                 & 35-44            & 15.8\%         &  &  \\ \cline{2-3}
\multicolumn{1}{|l|}{}                                 & 45-54            & 12.6\%         &  &  \\ \cline{2-3}
\multicolumn{1}{|l|}{}                                 & 55-59            & 4.6\%          &  &  \\ \cline{2-3}
\multicolumn{1}{|l|}{}                                 & 60-64            & 3.9\%          &  &  \\ \cline{2-3}
\multicolumn{1}{|l|}{}                                 & 65-74            & 6.2\%          &  &  \\ \cline{2-3}
\multicolumn{1}{|l|}{}                                 & 77-84            & 4.0\%          &  &  \\ \cline{2-3}
\multicolumn{1}{|l|}{}                                 & \textgreater{}85 & 1.5\%          &  &  \\ \cline{1-3}
\multicolumn{1}{|l|}{\multirow{2}{*}{\textbf{Gender}}} & Male             & 47.38\%        &  &  \\ \cline{2-3}
\multicolumn{1}{|l|}{}                                 & Female           & 52.62\%        &  &  \\ \cline{1-3}
\end{tabular}
\caption{Individual Characteristics}
\label{tbl:IndivChar}
\end{table}

\subsection{Geographic locations}

The spatial layout included ten groups of adjacent Neighborhood Tabulation Areas (NTA's) in Manhattan \cite{NYCData}. The reason for grouping NTAs stemmed from the granularity of the data that is available for the number of business in NYC \cite{NYCHouseholds}. 

NTAs in our data were virtually positioned in our spatial layout according to its central longitude and latitude. Each NTA included multiple locations of five types: households, workplaces, schools, stations, and supermarkets, which we located uniformly around the center. The number of households and workplaces in each NTAs were taken from \cite{NYCData} and \cite{NYCHouseholds}, respectively. A total of 1700 schools \cite{NYCEducation}, and 470 stations \cite{NTA}, were evenly divided  across NTAs. The number of supermarkets where computed based on one supermarket per each 100000 inhabitants statistics \cite{hoertel2020facing}. Other location types, such as public parks, shops and theathers, were modeled via abrupt disease infection, as we explain in Appendix B.

Table \ref{tbl:Locs} provides the list of NTAs, their coordinates, and the location types and numbers per NTA. 

\begin{table}[h]
\centering
\begin{tabular}{llllllll}
\hline
\multicolumn{1}{|l|}{\textbf{NTA}}                                                                                  & \multicolumn{1}{l|}{\textbf{\RotText{longitude}}} & \multicolumn{1}{l|}{\textbf{\RotText{latitude}}} & \multicolumn{1}{l|}{\textbf{\RotText{households}}} & \multicolumn{1}{l|}{\textbf{\RotText{schools}}} & \multicolumn{1}{l|}{\textbf{\RotText{stations}}} & \multicolumn{1}{l|}{\textbf{\RotText{supermarkets}}} & \multicolumn{1}{l|}{\textbf{\RotText{workplaces}}} \\ \hline
\multicolumn{1}{|l|}{\begin{tabular}[c]{@{}l@{}}Battery Park City, \\ Greenwich Village \& \\ Soho\end{tabular}}    & \multicolumn{1}{l|}{40.86183}           & \multicolumn{1}{l|}{73.92363}          & \multicolumn{1}{l|}{61672}                  & \multicolumn{1}{l|}{170}                 & \multicolumn{1}{l|}{47}                   & \multicolumn{1}{l|}{35}                       & \multicolumn{1}{l|}{18626}                  \\ \hline
\multicolumn{1}{|l|}{Central Harlem}                                                                                & \multicolumn{1}{l|}{40.8221}            & \multicolumn{1}{l|}{73.92363}          & \multicolumn{1}{l|}{48680}                  & \multicolumn{1}{l|}{170}                 & \multicolumn{1}{l|}{47}                   & \multicolumn{1}{l|}{28}                       & \multicolumn{1}{l|}{1270}                   \\ \hline
\multicolumn{1}{|l|}{\begin{tabular}[c]{@{}l@{}}Chelsea, Clinton \& \\ Midtown Business District\end{tabular}}      & \multicolumn{1}{l|}{40.8089}            & \multicolumn{1}{l|}{73.9482}           & \multicolumn{1}{l|}{59821}                  & \multicolumn{1}{l|}{170}                 & \multicolumn{1}{l|}{47}                   & \multicolumn{1}{l|}{34}                       & \multicolumn{1}{l|}{32137}                  \\ \hline
\multicolumn{1}{|l|}{\begin{tabular}[c]{@{}l@{}}Chinatown \& \\ Lower East Side\end{tabular}}                       & \multicolumn{1}{l|}{40.7957}            & \multicolumn{1}{l|}{73.9389}           & \multicolumn{1}{l|}{67708}                  & \multicolumn{1}{l|}{170}                 & \multicolumn{1}{l|}{47}                   & \multicolumn{1}{l|}{39}                       & \multicolumn{1}{l|}{8573}                   \\ \hline
\multicolumn{1}{|l|}{East Harlem}                                                                                   & \multicolumn{1}{l|}{40.7736}            & \multicolumn{1}{l|}{73.9566}           & \multicolumn{1}{l|}{48174}                  & \multicolumn{1}{l|}{170}                 & \multicolumn{1}{l|}{47}                   & \multicolumn{1}{l|}{28}                       & \multicolumn{1}{l|}{1721}                   \\ \hline
\multicolumn{1}{|l|}{\begin{tabular}[c]{@{}l@{}}Hamilton Heights, \\ Manhattanville \& \\ West Harlem\end{tabular}} & \multicolumn{1}{l|}{40.787}             & \multicolumn{1}{l|}{73.9754}           & \multicolumn{1}{l|}{51826}                  & \multicolumn{1}{l|}{170}                 & \multicolumn{1}{l|}{47}                   & \multicolumn{1}{l|}{30}                       & \multicolumn{1}{l|}{1962}                   \\ \hline
\multicolumn{1}{|l|}{\begin{tabular}[c]{@{}l@{}}Murray Hill, \\ Gramercy \& \\ Stuyvesant Town\end{tabular}}        & \multicolumn{1}{l|}{40.75507}           & \multicolumn{1}{l|}{73.9924}           & \multicolumn{1}{l|}{62339}                  & \multicolumn{1}{l|}{170}                 & \multicolumn{1}{l|}{47}                   & \multicolumn{1}{l|}{36}                       & \multicolumn{1}{l|}{22156}                  \\ \hline
\multicolumn{1}{|l|}{Upper East Side}                                                                               & \multicolumn{1}{l|}{40.7388}            & \multicolumn{1}{l|}{73.97933}          & \multicolumn{1}{l|}{91671}                  & \multicolumn{1}{l|}{170}                 & \multicolumn{1}{l|}{47}                   & \multicolumn{1}{l|}{53}                       & \multicolumn{1}{l|}{8896}                   \\ \hline
\multicolumn{1}{|l|}{\begin{tabular}[c]{@{}l@{}}Upper West Side \& \\ West Side\end{tabular}}                       & \multicolumn{1}{l|}{40.7154}            & \multicolumn{1}{l|}{73.99065}          & \multicolumn{1}{l|}{80736}                  & \multicolumn{1}{l|}{170}                 & \multicolumn{1}{l|}{47}                   & \multicolumn{1}{l|}{46}                       & \multicolumn{1}{l|}{6406}                   \\ \hline
\multicolumn{1}{|l|}{\begin{tabular}[c]{@{}l@{}}Washington Heights, \\ Inwood \& \\ Marble Hill\end{tabular}}       & \multicolumn{1}{l|}{40.72283}           & \multicolumn{1}{l|}{74.00717}          & \multicolumn{1}{l|}{81367}                  & \multicolumn{1}{l|}{170}                 & \multicolumn{1}{l|}{47}                   & \multicolumn{1}{l|}{47}                       & \multicolumn{1}{l|}{3049}                   \\ \hline              
\end{tabular}
\caption{Summary of NTAs in Manhattan}
\label{tbl:Locs}
\end{table}

\subsection{Social Interactions}
Agents in the simulation were assigned to locations in the virtual grid. The visit frequency to assigned location depends on the location type, as we describe below. Interactions between agents occurred via visits to the same location. 

The agents assignment to locations, and the meeting probability at each location type is as follows. 

\textbf{Households}. Agents were assigned to households randomly according to the family structure in NYC \cite{USCensus}. Daily meeting probability within each household was equal to 1. 

\textbf{Schools}. kids of age 6-18 were assigned to the closest school to their household. Within each schools, kids were clustered into classes according to their age group, and average class size of 26 \cite{NYCEducation}. Meeting probability within class was assumed to be 1 and 0 otherwise.

\textbf{Workplaces}. Adults of age 20-60 were assigned to Workplaces according to the NYC employment rate (95.9\%, \cite{Labor}). Assignment was not geo-dependent. Following \cite{hoertel2020facing}, 89\% of the workplaces were considered small. In these places, workers were clustered into working groups of three colleagues on average. Large workplaces accounted for 11\% of all workplaces, with workers grouped into groups of 11 workers on average. Meeting probability with colleagues was assumed to be 1, and 0 with other workers. 

\textbf{Supermarkets}. We assumed working agents go to supermarkets that are closest to their household and workplace. Non-working adult agents were assumed to choose the supermarket that is closest to their household. Visits to a supermarket occurred on average twice a week (assumption). Meeting probability of two agents that are assigned to the same supermarket was computed such that each agents interacts with an average of 30 other agents in a visit. 

\textbf{Stations}. Working agents were assumed to use the subway stations that are closest to their household and workplace. Meeting probability of two agents that go to the same station is computed such that each agents interacts with an average of 10 other agents in a ride. 

Table \ref{tbl:interactions} summarizes the social interactions.

\begin{table}[]
\begin{tabular}{lllcccll}
\cline{1-6}
\multicolumn{1}{|l|}{\textbf{Location type}}                                                                      & \multicolumn{2}{l|}{\textbf{Cluster by}}                                              & \multicolumn{1}{l|}{\textbf{\begin{tabular}[c]{@{}l@{}}Average   \\ interactions\end{tabular}}}            & \multicolumn{1}{l|}{\textbf{\begin{tabular}[c]{@{}l@{}}Visit \\ probability\end{tabular}}} & \multicolumn{1}{l|}{\textbf{\begin{tabular}[c]{@{}l@{}}Meeting  \\ probability\end{tabular}}} & \textbf{} & \textbf{} \\ \cline{1-6}
\multicolumn{1}{|l|}{\multirow{5}{*}{\textbf{Household}}}                                                         & \multicolumn{2}{l|}{Family structure}                                                 & \multicolumn{1}{l|}{}                                                                                      & \multicolumn{1}{c|}{\multirow{5}{*}{1}}                                                    & \multicolumn{1}{c|}{\multirow{5}{*}{1}}                                                       &           &           \\ \cline{2-4}
\multicolumn{1}{|l|}{}                                                                                            & \multicolumn{1}{l|}{\textit{Singles}}                  & \multicolumn{1}{l|}{32.20\%} & \multicolumn{1}{c|}{\multirow{4}{*}{\textit{\begin{tabular}[c]{@{}c@{}}entire \\ household\end{tabular}}}} & \multicolumn{1}{c|}{}                                                                      & \multicolumn{1}{c|}{}                                                                         &           &           \\ \cline{2-3}
\multicolumn{1}{|l|}{}                                                                                            & \multicolumn{1}{l|}{\textit{Couples with children}}    & \multicolumn{1}{l|}{41.20\%} & \multicolumn{1}{c|}{}                                                                                      & \multicolumn{1}{c|}{}                                                                      & \multicolumn{1}{c|}{}                                                                         &           &           \\ \cline{2-3}
\multicolumn{1}{|l|}{}                                                                                            & \multicolumn{1}{l|}{\textit{Couples without children}} & \multicolumn{1}{l|}{15.80\%} & \multicolumn{1}{c|}{}                                                                                      & \multicolumn{1}{c|}{}                                                                      & \multicolumn{1}{c|}{}                                                                         &           &           \\ \cline{2-3}
\multicolumn{1}{|l|}{}                                                                                            & \multicolumn{1}{l|}{\textit{Singles with children}}    & \multicolumn{1}{l|}{10.80\%} & \multicolumn{1}{c|}{}                                                                                      & \multicolumn{1}{c|}{}                                                                      & \multicolumn{1}{c|}{}                                                                         &           &           \\ \cline{1-6}
\multicolumn{1}{|l|}{\textbf{School  (ages 6-18)}}                                                                & \multicolumn{2}{l|}{Age groups, location of household}                                & \multicolumn{1}{c|}{26}                                                                                    & \multicolumn{1}{c|}{1}                                                                     & \multicolumn{1}{c|}{1}                                                                        &           &           \\ \cline{1-6}
\multicolumn{1}{|l|}{\multirow{3}{*}{\textbf{\begin{tabular}[c]{@{}l@{}}Wokplace  \\ (ages 18-65)\end{tabular}}}} & \multicolumn{2}{l|}{Workplace size}                                                   & \multicolumn{1}{c|}{}                                                                                      & \multicolumn{1}{c|}{\multirow{3}{*}{1}}                                                    & \multicolumn{1}{c|}{\multirow{3}{*}{1}}                                                       &           &           \\ \cline{2-4}
\multicolumn{1}{|l|}{}                                                                                            & \multicolumn{1}{l|}{\textit{Small}}                    & \multicolumn{1}{l|}{89\%}    & \multicolumn{1}{c|}{3}                                                                                     & \multicolumn{1}{c|}{}                                                                      & \multicolumn{1}{c|}{}                                                                         &           &           \\ \cline{2-4}
\multicolumn{1}{|l|}{}                                                                                            & \multicolumn{1}{l|}{\textit{Large}}                    & \multicolumn{1}{l|}{11\%}    & \multicolumn{1}{c|}{11}                                                                                    & \multicolumn{1}{c|}{}                                                                      & \multicolumn{1}{c|}{}                                                                         &           &           \\ \cline{1-6}
\multicolumn{1}{|l|}{\multirow{2}{*}{\textbf{\begin{tabular}[c]{@{}l@{}}Supermarket \\ (adults)\end{tabular}}}}   & \multicolumn{2}{l|}{Location of household  and workplace}                             & \multicolumn{1}{c|}{\begin{tabular}[c]{@{}c@{}}30  random \\ contacts\end{tabular}}                        & \multicolumn{1}{c|}{2/7}                                                                   & \multicolumn{1}{c|}{6.70E-04}                                                                 &           &           \\ \cline{2-6}
\multicolumn{1}{|l|}{}                                                                                            & \multicolumn{2}{l|}{Location of household  and workplace}                             & \multicolumn{1}{c|}{\begin{tabular}[c]{@{}c@{}}10  random \\ contacts\end{tabular}}                        & \multicolumn{1}{c|}{1}                                                                     & \multicolumn{1}{c|}{4.09E-03}                                                                 &           &           \\ \cline{1-6}
                                                                                                                  &                                                        &                              & \multicolumn{1}{l}{}                                                                                       & \multicolumn{1}{l}{}                                                                       & \multicolumn{1}{l}{}                                                                          &           &           \\
                                                                                                                  &                                                        &                              & \multicolumn{1}{l}{}                                                                                       & \multicolumn{1}{l}{}                                                                       & \multicolumn{1}{l}{}                                                                          &           &           \\
                                                                                                                  &                                                        &                              & \multicolumn{1}{l}{}                                                                                       & \multicolumn{1}{l}{}                                                                       & \multicolumn{1}{l}{}                                                                          &           &           \\
                                                                                                                  &                                                        &                              & \multicolumn{1}{l}{}                                                                                       & \multicolumn{1}{l}{}                                                                       & \multicolumn{1}{l}{}                                                                          &           &          
\end{tabular}
\caption{Summary of Social Interactions}
\label{tbl:interactions}
\end{table}

\end{document}